\pdfoutput=1

\documentclass[11pt]{article}

\usepackage[final]{acl}

\usepackage{times}
\usepackage{latexsym}

\usepackage[T1]{fontenc}

\usepackage[utf8]{inputenc}

\usepackage{microtype}

\usepackage{inconsolata}

\usepackage{graphicx}

\title{How much do LLMs learn from negative examples?}

\author{Shadi Hamdan \hspace*{7mm} Deniz Yuret\\
    KUIS AI Center, Koç University \\
  \texttt{shamdan17,dyuret@ku.edu.tr}\\
  }
\usepackage{pgfplots}
\pgfplotsset{width=70mm,compat=1.9}

\usepackage{amsmath}

\begin{document}
\maketitle
\begin{abstract}
Large language models (LLMs) undergo a three-phase training process: unsupervised pre-training, supervised fine-tuning (SFT), and learning from human feedback (RLHF/DPO). Notably, it is during the final phase that these models are exposed to negative examples—incorrect, rejected, or suboptimal responses to queries. This paper delves into the role of negative examples in the training of LLMs, using a likelihood-ratio (Likra) model on multiple-choice question answering benchmarks to precisely manage the influence and the volume of negative examples. 
Our findings reveal three key insights: (1) During a critical phase in training, Likra with negative examples demonstrates a significantly larger  improvement per training example compared to SFT using only positive examples. This leads to a sharp jump in the learning curve for Likra unlike the smooth and gradual improvement of SFT; (2) negative examples that are plausible but incorrect (near-misses) exert a greater influence; and (3) while training with positive examples fails to significantly decrease the likelihood of plausible but incorrect answers, training with negative examples more accurately identifies them. These results indicate a potentially significant role for negative examples in improving accuracy and reducing hallucinations for LLMs.
\end{abstract}

\section{Introduction}

Large language models are typically pre-trained on next word prediction over large collections of text, then fine-tuned on desired responses to user prompts. They only encounter negative examples in the final stage of their training in the form of false, undesirable, unsafe, or low-quality outputs. Techniques like reinforcement learning from human feedback (RLHF) \cite{ouyang2022training} or direct preference optimization (DPO) \cite{rafailov2024direct} are used to train from human preference data that includes  such negative examples. RLHF learns a reward function that imitates user preferences and uses reinforcement learning to align the generation process with these preferences. DPO bypasses reward learning and directly trains the model on pairs of good and bad outputs in the training set.

In this paper we focus on the contribution of negative examples for language model training and find that their impact is both qualitatively and quantitatively different from positive examples. Specifically we demonstrate that (1) during a critical phase in training, each additional negative example can improve the accuracy of a model $10\times$ more than each additional positive example resulting in a sharp jump in the learning curve, (2) near-miss negative examples, i.e. plausible sounding but incorrect outputs, are a lot more effective in training, and (3) models exposed to negative examples are a lot better at differentiating correct answers from plausible but incorrect ones at inference time.

\begin{figure*}[t]
\begin{center}
\begin{tikzpicture}
\begin{semilogxaxis}[
    width=100mm,
    height=70mm,
    title={SFT vs Likra learning curves},
    xlabel={training examples},
    ylabel={ARC-Challenge acc-norm},
    xmin=0.5, xmax=10000,
    ymin=0.58, ymax=0.83,
    legend pos=north west,
    ymajorgrids=true,
    grid style=dashed,
]
\addplot+[
    error bars/.cd, y dir=both, y explicit,
    ]
    table [x=x,y=y,y error=std] {
        x       y       std
        1   0.5998  0.0000
        16	0.6143	0.0000
        32	0.6160	0.0000
        64	0.6135	0.0000
        128	0.6118	0.0000
        256	0.6246	0.0000
        512	0.6391	0.0000
        1024	0.6271	0.0000
        2048	0.6399	0.0000
        4096	0.6476	0.0000
        6615	0.6630	0.0000
    };
    \addplot+[
    mark=triangle*,
    error bars/.cd, y dir=both, y explicit,
    ]
    table [x=x,y=y,y error=std] {
        x       y       std
        16	0.3618	0.0000
        32	0.3993	0.0000
        64	0.6314	0.0000
        128	0.7637	0.0000
        256	0.7986	0.0000
        512	0.7995	0.0000
        1024	0.8012	0.0000
        2048	0.8123	0.0000
        4096	0.8123	0.0000
        6615    0.8183	0.0000
    };
    \legend{
    \textsc{sft}, 
    \textsc{likra}
    }

\end{semilogxaxis}
\end{tikzpicture}
\end{center}
\caption{Comparison of learning curves for supervised fine-tuning (SFT) and likelihood-ratio
(Likra) models for the ARC-Challenge benchmark \cite{clark2018think} using Mistral-7B-v0.1 \cite{jiang2023mistral} as a base model. The SFT model is the result of regular supervised fine-tuning using only correct question-answer pairs.
The Likra model uses an equal number of incorrect question-answer pairs to train a negative head, uses the likelihood ratio of the SFT model and the negative head to decide on its answers.}
\label{fig:likra}
\end{figure*}
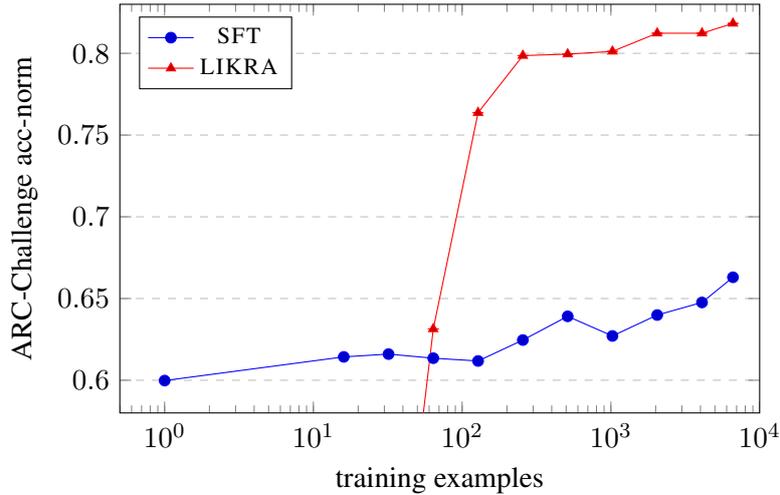

Negative examples can help guide the learning process by providing explicit information about what the model should avoid generating or classifying as positive. This approach can enhance the model's discriminative ability or refine its generative output. The use of negative examples in machine learning goes back to Patrick Winston's pioneering work on the importance of ``near-miss'' examples in concept learning \cite{Winston1970}. More recently, techniques such as hard negative mining \cite{4587597} and adversarial example generation \cite{szegedy2013intriguing} have used near-miss negative examples to measure and improve the robustness of discriminative models. The incorporation of negative examples can also make generative models more discerning and controlled, reducing the likelihood of generating undesirable outputs. Contrastive divergence \cite{hinton2002training}, auto-encoders \cite{hinton2006reducing} and generative adversarial networks \cite{goodfellow2014generative} learn by contrasting real examples with model-generated ones, unlikelihood training \cite{welleck2019neural} explicitly penalizes models for generating certain types of undesirable outputs, and noise-contrastive estimation \cite{gutmann2010noise} tackles modeling intractable distributions using negative ``noise'' data. 

We use a likelihood-ratio model to study the impact of negative examples. Likelihood-ratio models have long been used in classification tasks where the likelihoods from multiple models are compared and used as a decision criterion or to identify out of distribution data: Naive Bayes classifiers, Gaussian Mixture Models are basic examples. In generative modeling, noise-contrastive estimation contrasts the target distribution with a ``noise'' distribution to construct a tractable loss function and unlikelihood training combines two likelihood terms to prevent generation of repetitive and dull text.

We chose to train two independent models (in practice two LoRA adapters on a single foundation model), one on positive examples the other on negative examples, and use their likelihood-ratio (hence the name Likra) during inference to isolate and quantify the impact of negative examples. This allows us to vary the number of positive and negative examples independently during training and control the relative weights of the two models during inference (unlike e.g. a single DPO model). The downside is that the resulting Likra model cannot be easily used for generation, so all our testing is done on multiple-choice benchmarks\footnote{The code to replicate our experiments is available at \url{https://github.com/Shamdan17/likra}}.

Figure~\ref{fig:likra} illustrates a key result of our work: when trained on an equal number of positive and negative examples for a multiple-choice task, Likra exhibits a sharp, step-function-like increase in accuracy after a few hundred examples. This behavior contrasts sharply with the smooth and gradual learning curve observed in supervised fine-tuning. Details for this result are provided in Section~\ref{sec:exp}.

After formally describing Likra in Section~\ref{sec:likra}, and presenting our main results in Section~\ref{sec:exp}, we run a series of ablation experiments to help understand this strange contribution of negative examples in Section~\ref{sec:analysis}. Section~\ref{sec:discussion} summarizes our findings.

\section{Likra: the likelihood-ratio model} \label{sec:likra}

In our experiments we use a likelihood-ratio language model (Likra) to isolate and compare the contributions of positive and negative examples during training. The model consists of two heads: the positive head is trained on correct question-answer pairs, and the negative head is trained on incorrect/undesirable question-answer pairs. Each head is trained to maximize the conditional log-likelihood of the corresponding answer given the question:
\begin{align}
    \mathcal{L}_{\textsc{mle}}(p_\theta, \mathcal{D}) &= \sum_{q,a\in\mathcal{D}} \sum_{t=1}^{|a|} \log p_\theta(a_t | q, a_{<t})
\end{align}
where $\mathcal{L_\textsc{mle}}$ is the likelihood, $\theta$ represents model parameters, $\mathcal{D}$ is the training data, $q, a$ are question-answer pairs (correct answer for the positive head, incorrect answer for the negative head), $a_t$ is the $t$'th token of the answer.

Each head is independently trained starting from a base pre-trained language model, the positive head giving higher likelihood $\mathcal{L}^+$ to correct answers and the negative head giving higher likelihood $\mathcal{L}^-$ to incorrect answers. During inference time we use the log likelihood ratio $\mathcal{L}^+ - \mathcal{L}^-$ to score answer candidates.

\section{Experiments} \label{sec:exp}

In this section we lay out the empirical evidence for the main thesis of this paper: starting from a pre-trained language model, during a critical phase of the training, negative examples (questions paired with incorrect answers) have a significantly larger impact on accuracy than  positive examples (questions paired with correct answers), which leads to a sharp jump in the learning curve unlike the smooth and gradual improvement of SFT.

We start with a fairly standard supervised fine-tuning example where a pre-trained base language model is fine-tuned with correct question-answer pairs from the training set of a standard multiple-choice benchmark. We call this the SFT model.
We then start with the same base model / dataset and train a negative model by fine-tuning with incorrect question-answer pairs. The Likra model chooses answers based on the likelihood ratio of these two models and demonstrates the step-function-like jump in accuracy in its training curve and significantly outperforms positive-example-only trained SFT model.

For brevity, the discussion below uses the results from the Mistral-7B-v0.1 \cite{jiang2023mistral} base model and the ARC \cite{clark2018think} benchmark unless otherwise noted. Experiments with other base models and multiple-choice benchmarks show similar results and are summarized at the end of the section.

\subsection{Supervised fine-tuning}

In this section we start with an example of supervised fine-tuning (training a base model with correct question-answer pairs) resulting in modest gains in accuracy on a multiple-choice benchmark. 

To construct the training set we used the AI2 Reasoning Challenge (ARC) benchmark, a set of grade-school level multiple-choice science questions \cite{clark2018think} such as: 

\begingroup\ttfamily\noindent \\
What can a flower become?\\
(A) a fruit\\
(B) a leaf  \\
(C) a stem  \\
(D) a branch\\
\\
Which substance is a compound?\\
(A) sodium  \\
(B) chlorine  \\
(C) table salt  \\
(D) salt water \\
\endgroup

We used the \textsc{lm-evaluation-harness} \cite{eval-harness} for evaluation, which prepends 25 few-shot examples (random correct question-answer pairs) to each test question and compares the per-character likelihoods of different answer choices. We excluded the 1172 questions used by \textsc{lm-evaluation-harness} from our training set and we paired the remaining 6615 questions with their correct answers for supervised fine-tuning:

\begingroup\ttfamily\noindent\\
Question: What can a flower become?\\
Answer: a fruit\\
\\
Question: Which substance is a compound?\\
Answer: table salt\\
\endgroup

We used Mistral-7B-v0.1 as a base model \cite{jiang2023mistral} for supervised fine-tuning. The training was performed with zero-shot examples (no extra questions in the context) optimizing the likelihood of the correct answer conditional on the question (the question logits were ignored). 
We trained a LoRA adapter \cite{hu2021lora} only for a single epoch (more epochs did not help) using batch size 8 and the Adam optimizer \cite{kingma2014adam} with learning rate $10^{-4}$.

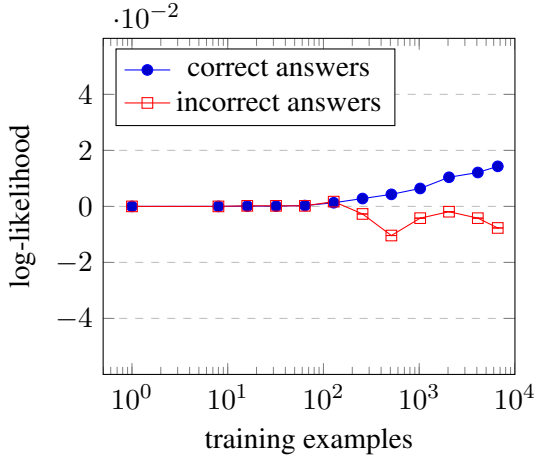
\begin{figure}[h]
\begin{center}
\begin{tikzpicture}
\begin{semilogxaxis}[
    xlabel={training examples},
    ylabel={log-likelihood},
    xmin=0.5, xmax=10000,
    ymin=-0.06, ymax=0.06,
    ytick={-0.04,-0.02,0,0.02,0.04},
    legend pos=north west,
    ymajorgrids=true,
    grid style=dashed,
]
\addplot+[
        error bars/.cd, y dir=both, y explicit,
        ]
table [x=x,y=y,y error=std] { 
	x       y       std 
        1 0.0000 0.0000
	8 0.0000 0.0000
	16 0.0001 0.0000
	32 0.0001 0.0000
	64 0.0003 0.0000
	128 0.0013 0.0000
	256 0.0028 0.0000
	512 0.0043 0.0000
	1024 0.0064 0.0000
	2048 0.0104 0.0000
	4096 0.0121 0.0000
	6615 0.0143 0.0000
};

\addplot+[
        mark=square,
        error bars/.cd, y dir=both, y explicit,
        ]
table [x=x,y=y,y error=std] { 
	x       y       std
        1 0.0000 0.0000
	8 0.0000 0.0000
	16 0.0002 0.0000
	32 0.0002 0.0000
	64 0.0002 0.0000
	128 0.0017 0.0000
	256 -0.0027 0.0000
	512 -0.0104 0.0000
	1024 -0.0042 0.0000
	2048 -0.0019 0.0000
	4096 -0.0042 0.0000
	6615 -0.0077 0.0000
};

    \legend{correct answers, incorrect answers}
\end{semilogxaxis}
\end{tikzpicture}
\end{center}
\caption{Likelihood of the correct answer vs the most likely incorrect answer during training.}
\label{fig:sft}
\end{figure}

The SFT learning curve in Figure~\ref{fig:likra} was obtained by training the base model on random samples with 0 to 6615 (full dataset) correct question-answer pairs from the training set. It shows a modest increase in accuracy (60\% to 66\%) as expected using in-domain training examples. Figure~\ref{fig:sft} compares the average per-character likelihood of the correct answers with the most likely incorrect answer throughout training. 
Even though the correct answers seem to increase in likelihood, the incorrect answers do not seem to be sharply distinguished. In Section~\ref{sec:probmass} we will look at the allocation of the model's probability mass in more detail and show that a model trained with negative examples can distinguish correct from incorrect more sharply.
In the next section we look at how the model can learn more from incorrect answers by using them as negative examples.

\subsection{Likra with negative examples}

In this section we show that negative examples (incorrect question-answer pairs) have a {\em significantly} larger effect on final accuracy compared to positive examples (correct question-answer pairs). 
We experimented with a Likra model rather than RLHF or DPO  because this made it easier to isolate and control the contribution of negative examples. The results show that doubling the number of positive examples increases the accuracy of the SFT model by less than 1\%, whereas doubling the number of negative examples can increase the accuracy of the Likra model by more than 10\% during the critical phase of training. 

To generate a training set of negative examples we paired each question in the ARC training set with an incorrect answer chosen randomly from the multiple-choice options, e.g.:

\begingroup\ttfamily\noindent\\
Question: What can a flower become?\\
Answer: a leaf\\
\\
Question: Which substance is a compound?\\
Answer: salt water\\
\endgroup

We used Mistral-7B-v0.1 as a base model for Likra. To train the negative head we followed a procedure similar to SFT training except for using negative examples. 

Figure~\ref{fig:likra} compares the SFT model which has been fine-tuned with only positive examples with a Likra model that uses the same SFT model as its positive head in addition to a negative head fine-tuned with incorrect question-answer pairs. We observe that starting from a well pre-trained base model, the contribution of each negative example to the final accuracy far exceeds the contribution of each positive example in the critical training phase at 64-256 examples. Increasing the number of negative training examples from 64 to 128 (only 8 extra updates with a batch size of 8) adds nearly 15\% accuracy, whereas the SFT model averages less than 1\% improvement per doubling of positive examples. 

\emph{It is unlikely for the model to learn much new information from just a few wrong answers, instead the negative examples seem to quite rapidly unlock latent knowledge that already exists in the pretrained model.}

\subsection{Other base models and benchmarks}

\begin{table}[h]
\centering
\caption{Final accuracy on ARC-Challenge and HellaSwag. No superscript in the model name indicates the base model, $^+$ indicates supervised fine-tuning with positive examples, $^-$ indicates the Likra model trained with both positive and negative examples.}
\label{tab:results}
\begin{tabular}{|lll|} \hline
Model & ARC & HellaSwag \\
\hline
Mistral-7B-v0.1 & .5998 & .8323 \\
Mistral-7B-v0.1$^+$ & .6630 & .8468 \\
Mistral-7B-v0.1$^-$ & {\bf .8123} & {\bf .9633} \\
Mistral-7B-Instruct-v0.3 & .6365 & .8463 \\
Mistral-7B-Instruct-v0.3$^+$ & .6408 & .8360 \\
Mistral-7B-Instruct-v0.3$^-$ & {\bf .8063} & {\bf .9569} \\
Llama-3.2-3B-Instruct & .5222 & .7312 \\
Llama-3.2-3B-Instruct$^+$ & .5486 & .7254 \\
Llama-3.2-3B-Instruct$^-$ & {\bf .7321} & {\bf .9071} \\
\hline\end{tabular}
\end{table}

In order to test the generality of our results we experimented with two benchmarks and three models. Table~\ref{tab:results} summarizes the results. 

ARC-Challenge benchmark \cite{clark2018think} is a set of grade-school level English multiple-choice science questions with 6615 training and 1172 test instances. The fine-tuning for ARC takes around 5 minutes on 1$\times$A40 and the evaluation takes around 15 minutes on 8$\times$A40. HellaSwag \cite{zellers2019hellaswag} is a set of multiple-choice English text completion tasks based on video descriptions and how-to manuals with 39905 training and 10042 test instances. The fine-tuning for Hellaswag takes around 1 hour on 1$\times$A40 and the evaluation takes around 30 minutes on 8$\times$A40. Other than both being multiple-choice tasks, ARC and Hellaswag require quite different types of knowledge demonstrating some domain independence for our findings. Mistral \cite{jiang2023mistral} and Llama \cite{grattafiori2024llama3herdmodels} are open source foundation models\footnote{These resources have the following licenses: ARC-Challenge: CC BY-SA 4.0; HellaSwag: MIT; Mistral: Apache 2.0; Llama: Llama 3.2 License. All models allow usage and modification, and none of the datasets contain personal information that require anonymization.}. In each case we see a large jump in accuracy with the Likra model.

\section{Analysis} \label{sec:analysis}

In this section we probe the training process more deeply to understand the role of negative examples in boosting model accuracy. First we change the ratio of the positive and negative examples during training and the weight of the positive and negative heads during inference. The results ensure us that negative examples have a significantly stronger effect compared to positive examples. Then we categorize negative examples as incorrect, irrelevant, or unrelated and train different models with each category. The results show that the more plausible incorrect answers increase model accuracy the most. Finally we look at how likelihoods of different answer types (correct, incorrect, irrelevant, unrelated) evolve during the training process. The results show that the negative head learns to sharply distinguish plausible but incorrect answers from correct ones, whereas the positive head assigns them closer likelihoods.


\subsection{Do we even need positive examples?}

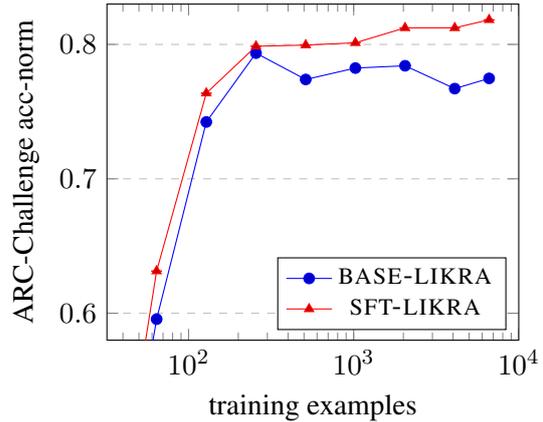
\begin{figure}[h]
\begin{center}
\begin{tikzpicture}
\begin{semilogxaxis}[
    xlabel={training examples},
    ylabel={ARC-Challenge acc-norm},
    xmin=32, xmax=10000,
    ymin=0.58, ymax=0.83,
    legend pos=south east,
    ymajorgrids=true,
    grid style=dashed,
]
    \addplot+[
    error bars/.cd, y dir=both, y explicit,
    ]
    table [x=x,y=y,y error=std] {
        x       y       std
        16	0.3712	0.0000
        32	0.3805	0.0000
        64	0.5956	0.0000
        128	0.7423	0.0000
        256	0.7935	0.0000
        512	0.7739	0.0000
        1024	0.7824	0.0000
        2048	0.7841	0.0000
        4096	0.7671	0.0000
        6615	0.7747	0.0000
    };
    \addplot+[
    mark=triangle*,
    error bars/.cd, y dir=both, y explicit,
    ]
    table [x=x,y=y,y error=std] {
        x       y       std
        16	0.3618	0.0000
        32	0.3993	0.0000
        64	0.6314	0.0000
        128	0.7637	0.0000
        256	0.7986	0.0000
        512	0.7995	0.0000
        1024	0.8012	0.0000
        2048	0.8123	0.0000
        4096	0.8123	0.0000
        6615    0.8183	0.0000
    };
    \legend{
    \textsc{base-likra}, 
    \textsc{sft-likra}, 
    }

\end{semilogxaxis}
\end{tikzpicture}
\end{center}
\caption{Likra with (SFT) and without (Base) positive examples.}
\label{fig:nopositive}
\end{figure}

The Likra model allows us to vary the number of positive and negative examples during training independently.
Given the large impact of negative examples on model accuracy, we asked if Likra would work without positive examples. In Figure~\ref{fig:nopositive} SFT-Likra uses the likelihood ratio of the SFT model and the negative model (same as Figure~\ref{fig:likra}), Base-Likra uses the likelihood ratio of the base model and the negative model. Effectively Base-Likra only uses negative examples for fine-tuning. The resulting learning curve of the two Likra models are fairly similar: they both have the step-function like accuracy increase at a few hundred examples and they both significantly outperform SFT.

\subsection{Varying the weight of the negative head}

\begin{figure}[h]
\begin{center}
\begin{tikzpicture}
\begin{axis}[
    xlabel={weight of the negative head},
    ylabel={ARC-Challenge acc-norm},
    xmin=-0.2, xmax=1.2,
    ymin=0.58, ymax=0.85,
    legend pos=north west,
    ymajorgrids=true,
    grid style=dashed,
]
\addplot+[
    error bars/.cd, y dir=both, y explicit,
    ]
    table [x=x,y=y,y error=std] {
        x       y       std
        0       0.5998  0.0000
        0.125	0.6852	0.0000	
        0.25	0.7073	0.0000	
        0.5	0.7491	0.0000	
        0.75	0.7986	0.0000	
        0.9    0.8191	0.0000
        1.0	0.8183	0.0000	
        1.1    0.8072	0.0000
        1.25	0.7867	0.0000	
        1.5	0.6945	0.0000	
        1.75	0.6126	0.0000	
        2.0	0.5580	0.0000	
    };
\end{axis}
\end{tikzpicture}
\end{center}
\caption{Changing the weight of the negative head.}
\label{fig:negativeweight}
\end{figure}
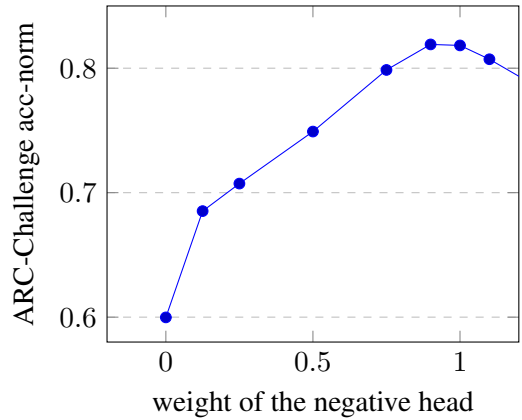

The Likra model allows us to vary the relative weight of the positive head and the negative head during inference. Figure~\ref{fig:negativeweight} plots the result of changing the weight of the negative head during inference, i.e. score = $\mathcal{L}^+ - \mbox{weight}\times\mathcal{L}^-$. It shows that the accuracy increases as the weight of the negative head is increased and peaks around 0.9-1.0.

\subsection{Near-miss negative examples teach more}

The Likra model works by contrasting the conditional likelihood given to an answer by a pre-trained (or SFT trained) positive head and a negative head trained on incorrect question-answer pairs. The results so far demonstrated the importance of training on negative examples. In this section we ask whether the plausibility of these incorrect answers matter during training, i.e. can we construct negative examples by answering questions with irrelevant text, or do the false answers have to be plausible? We conclude that even though all negative training can be beneficial, the near-miss negative examples consisting of plausible sounding but incorrect question-answer pairs work best.

We generated three different training sets of negative answers by pairing each question in the ARC training set with a false answer chosen from (i) multiple-choice options for that question, (ii) random false answer for a different ARC question, and (iii) random false answer from an unrelated benchmark (we tried non-science-related tests from MMLU \cite{hendrycks2020measuring} and HellaSwag \cite{zellers2019hellaswag} with similar results). For example:

\begingroup
\ttfamily
\noindent
Question: Which substance is a compound?\\
Answer: salt water\\
(incorrect from the same question)\\
\\
Question: Which substance is a compound?\\
Answer: reduce the energy requirements \\
(irrelevant from the same test (ARC))\\
\\
Question: Which substance is a compound?\\
Answer: is playing the piano \\
(unrelated from another test (HellaSwag))
\endgroup

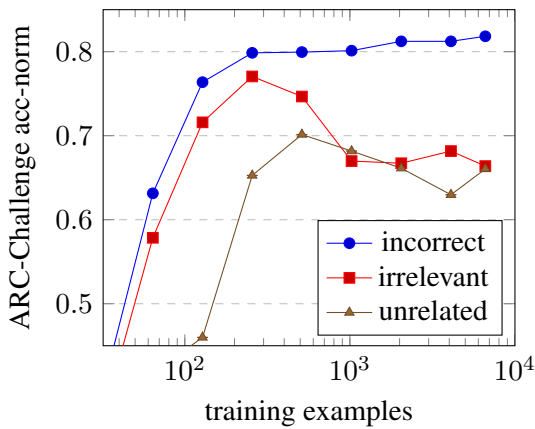
\begin{figure}[h]
\begin{center}
\begin{tikzpicture}
\begin{semilogxaxis}[
    xlabel={training examples},
    ylabel={ARC-Challenge acc-norm},
    xmin=32, xmax=10000,
    ymin=0.45, ymax=0.85,
    legend pos = south east,
    ymajorgrids=true,
    grid style=dashed,
]
    \addplot+[
        error bars/.cd, y dir=both, y explicit,
        ]
        table [x=x,y=y,y error=std] { 
            x       y       std
            16	0.3618	0.0000
            32	0.3993	0.0000
            64	0.6314	0.0000
            128	0.7637	0.0000
            256	0.7986	0.0000
            512	0.7995	0.0000
            1024	0.8012	0.0000
            2048	0.8123	0.0000
            4096	0.8123	0.0000
            6615    0.8183  0.0000
        };
    \addplot+[
        error bars/.cd, y dir=both, y explicit,
        ]
        table [x=x,y=y,y error=std] { 
             x       y       std
            32	0.3643	0.0000
            64	0.5785	0.0000
            128	0.7159	0.0000
            256	0.7705	0.0000
            512	0.7466	0.0000
            1024	0.6698	0.0000
            2048	0.6672	0.0000
            4096	0.6817	0.0000
            6615	0.6638	0.0000
        };
    \addplot+[ 
        mark=triangle*,
        error bars/.cd, y dir=both, y explicit,
        ]
        table [x=x,y=y,y error=std] { 
            x       y       std
            16	0.2867	0.000
            32	0.2756	0.000
            64	0.4172	0.000
            128	0.4599	0.000
            256	0.6527	0.000
            512	0.7014	0.000
            1024	0.6817	0.000
            2048	0.6613	0.000
            4096	0.6297	0.000
            6615	0.6604	0.000
        };
    \legend{incorrect, irrelevant, unrelated}
    \end{semilogxaxis}
\end{tikzpicture}
\end{center}
\caption{Training with different negative examples.}
\label{fig:negtypes}
\end{figure}

Figure~\ref{fig:negtypes} compares the performance of training the negative head with these three types of false answers (using the SFT model as the positive head). One surprising observation is that even when the false answers are completely unrelated random text (e.g. HellaSwag answers to ARC science questions) they are beneficial (the Likra model reaches 70\% accuracy outperforming the SFT model). Maybe less surprising is the finding that the more plausible false answers are the more beneficial they seem to be, as  suggested by Patrick Winston's pioneering observation of the importance of near-miss negative examples for learning \cite{Winston1970}.



\subsection{Where does the probability mass go?} \label{sec:probmass}

\begin{figure}[h]
\begin{center}
\begin{tikzpicture}
\begin{semilogxaxis}[
    title={Positive head},
    ylabel={log-likelihood},
    xmin=5, xmax=10000,  
    ymin=-0.06, ymax=0.06,
    ytick={-0.04,-0.02,0,0.02,0.04},
    legend pos=north west,
    ymajorgrids=true,
    grid style=dashed,
]
    \addplot+[
        error bars/.cd, y dir=both, y explicit,
        ]
table [x=x,y=y,y error=std] { 
	x       y       std 
	8 0.0000 0.0000
	16 0.0001 0.0000
	32 0.0001 0.0000
	64 0.0003 0.0000
	128 0.0013 0.0000
	256 0.0028 0.0000
	512 0.0043 0.0000
	1024 0.0064 0.0000
	2048 0.0104 0.0000
	4096 0.0121 0.0000
	6615 0.0143 0.0000
};
\addplot+[
        error bars/.cd, y dir=both, y explicit,
        ]
table [x=x,y=y,y error=std] { 
	x       y       std 
	8 0.0000 0.0000
	16 0.0002 0.0000
	32 0.0002 0.0000
	64 0.0002 0.0000
	128 0.0017 0.0000
	256 -0.0027 0.0000
	512 -0.0104 0.0000
	1024 -0.0042 0.0000
	2048 -0.0019 0.0000
	4096 -0.0042 0.0000
	6615 -0.0077 0.0000
};
\addplot+[
        mark=triangle*,
        error bars/.cd, y dir=both, y explicit,
        ]
table [x=x,y=y,y error=std] { 
	x       y       std 
	8 0.0000 0.0000
	16 -0.0004 0.0000
	32 -0.0003 0.0000
	64 -0.0019 0.0000
	128 -0.0062 0.0000
	256 -0.0233 0.0000
	512 -0.0320 0.0000
	1024 -0.0317 0.0000
	2048 -0.0337 0.0000
	4096 -0.0438 0.0000
	6615 -0.0447 0.0000
};
\addplot+[
        error bars/.cd, y dir=both, y explicit,
        ]
table [x=x,y=y,y error=std] { 
	x       y       std 
	8 0.0000 0.0000
	16 -0.0002 0.0000
	32 -0.0002 0.0000
	64 -0.0009 0.0000
	128 -0.0031 0.0000
	256 -0.0093 0.0000
	512 -0.0132 0.0000
	1024 -0.0152 0.0000
	2048 -0.0163 0.0000
	4096 -0.0218 0.0000
	6615 -0.0197 0.0000
};

    \legend{correct, incorrect, irrelevant, unrelated}
    \end{semilogxaxis}
\end{tikzpicture}
\begin{tikzpicture}
    \begin{semilogxaxis}[
    title={Negative head},
    xlabel={training examples},
    ylabel={log-likelihood},
    xmin=5, xmax=10000,
    ymin=-0.06, ymax=0.06,
    ytick={-0.04,-0.02,0,0.02,0.04},
    legend style={at={(1,1)},anchor=north west}, 
    ymajorgrids=true,
    grid style=dashed,
]
    \addplot+[
        error bars/.cd, y dir=both, y explicit,
        ]
table [x=x,y=y,y error=std] { 
	x       y       std 
	8 0.0000 0.0000
	16 0.0001 0.0000
	32 0.0001 0.0000
	64 0.0004 0.0000
	128 0.0009 0.0000
	256 -0.0180 0.0000
	512 -0.0234 0.0000
	1024 -0.0221 0.0000
	2048 -0.0191 0.0000
	4096 -0.0166 0.0000
	6615 -0.0183 0.0000
};
\addplot+[
        error bars/.cd, y dir=both, y explicit,
        ]
table [x=x,y=y,y error=std] { 
	x       y       std 
	8 0.0000 0.0000
	16 0.0007 0.0000
	32 0.0008 0.0000
	64 0.0029 0.0000
	128 0.0144 0.0000
	256 0.0414 0.0000
	512 0.0434 0.0000
	1024 0.0468 0.0000
	2048 0.0504 0.0000
	4096 0.0519 0.0000
	6615 0.0530 0.0000
};
\addplot+[
        mark=triangle*,
        error bars/.cd, y dir=both, y explicit,
        ]
table [x=x,y=y,y error=std] { 
	x       y       std 
	8 0.0000 0.0000
	16 0.0005 0.0000
	32 0.0005 0.0000
	64 0.0018 0.0000
	128 0.0085 0.0000
	256 0.0353 0.0000
	512 0.0354 0.0000
	1024 0.0301 0.0000
	2048 0.0262 0.0000
	4096 0.0129 0.0000
	6615 0.0192 0.0000
};
\addplot+[
        error bars/.cd, y dir=both, y explicit,
        ]
table [x=x,y=y,y error=std] { 
	x       y       std 
	8 0.0000 0.0000
	16 0.0000 0.0000
	32 0.0000 0.0000
	64 0.0001 0.0000
	128 0.0000 0.0000
	256 -0.0024 0.0000
	512 -0.0065 0.0000
	1024 -0.0103 0.0000
	2048 -0.0107 0.0000
	4096 -0.0162 0.0000
	6615 -0.0137 0.0000
};

    \end{semilogxaxis}
\end{tikzpicture}
\end{center}
\caption{Likelihood assigned to answer types.} 
\label{fig:probmass}
\end{figure}
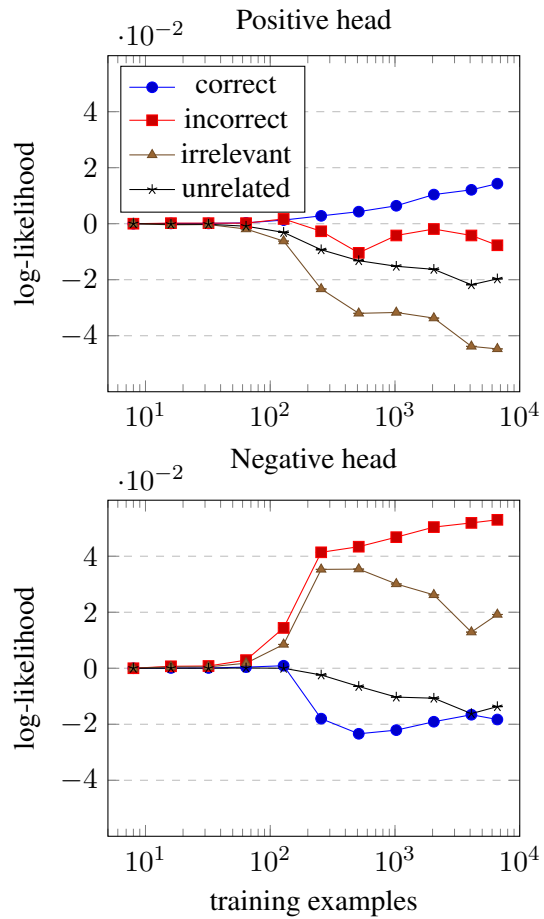

In this section we analyze how the probability mass shifts between different types of text during training to explain the performance of the Likra model. Our results show that the negative head of the Likra model becomes very good at identifying near-miss negative examples (plausible wrong answers) which compensates for the main weakness of the positive head.

To measure how the probability mass shifts between different regions of the text space during training, we paired {\em test set} questions with different types of answers (as opposed training set questions, like we did in the last section): the correct answer, the most likely incorrect answer, irrelevant text from the same benchmark (ARC), and unrelated text from a different benchmark (HellaSwag). We took the positive head (trained with correct answers), and the negative head (trained with random incorrect answers) at various points along their learning curve and evaluated their conditional likelihoods for these different types of text.

Figure~\ref{fig:probmass} shows the absolute change in per-character log-likelihood for different types of text. Text unrelated to the subject domain becomes less likely for both positive and negative heads. The positive head increases the likelihood of the correct answers as expected and decreases the likelihood of irrelevant/incorrect answers, however the likelihood of incorrect answers does not seem to decrease by much. The negative head increases the likelihood of incorrect answers significantly, irrelevant answers to a lesser extent, and decreases the likelihood of the correct answers. Even though the negative head has never seen a correct answer during fine-tuning it is able to distinguish them from plausible incorrect answers. 

When we take the difference of log likelihoods for inference, the biggest impact of the negative head turns out to be significantly decreasing the likelihood of incorrect answers. It seems that pre-trained language models can learn significantly more from plausible sounding incorrect answers, i.e. near-miss negative examples, than correct answers whose likelihoods are already relatively high in the base model.



\section{Discussion} \label{sec:discussion}
We still find it incredible that a few hundred negative examples improve the answer accuracy of a pre-trained language model significantly more than thousands of positive examples albeit in a restricted domain. It seems clear that {\em wrong} answers to a few hundred training questions cannot give the model much missing information about the test questions. Thus the knowledge to answer these test questions correctly must already reside in the pre-trained model but obfuscated by the probability mass given to other plausible sounding answers. Training with negative examples seems to flip a switch that causes the model to sharply distinguish factually accurate answers from plausible sounding ones. This supports a version of the ``Superficial Alignment Hypothesis'' \cite{zhou2024lima}: A model's knowledge and capabilities are learnt almost entirely during pretraining, and alignment teaches it not only format and style, but also preference for factual accuracy.

\section{Limitations} \label{sec:limitations}

We presented a method that improves the factual accuracy of large language models, however it does not guarantee that the resulting models will always generate or choose factually correct answers. The base models we use, as well as their fine-tuned versions may in some instances produce inaccurate, biased or other objectionable responses to user prompts. Our fine-tuning and evaluation only used English benchmarks. The Likra model specifically trains and uses a negative head specialized in recognizing factually inaccurate answers, however the two-head model structure makes it challenging to generate text. We suggest using Likra models to evaluate potential answers for accuracy, hallucination detection, or in multiple-choice testing scenarios. 

\bibliographystyle{acl_natbib}
\bibliography{main}


\end{document}